\documentclass[12pt,a4paper]{article}
\usepackage[utf8]{inputenc}
\usepackage[T1]{fontenc}
\usepackage[english]{babel}
\usepackage{amsmath}
\usepackage{amsfonts}
\usepackage{amssymb}
\usepackage{makeidx}
\usepackage{graphicx}
\usepackage{appendix}
\usepackage[utf8]{inputenc}
\usepackage{fancyhdr}
\usepackage{subcaption}

\usepackage{tcolorbox}
\usepackage{tabularx} 

\usepackage{authblk} 

\usepackage[symbol]{footmisc} 

\usepackage{tcolorbox}
\usepackage{lipsum}
\usepackage[colorlinks=True]{hyperref}
\usepackage[labelfont=bf]{caption}
\usepackage[margin=1in]{geometry}
\usepackage[numbers]{natbib}
\title{\textbf{KnowSLM: A framework for evaluation of small language models for knowledge augmentation and humanised conversations}}
\date{}
\begin{document}

\author[1]{Chitranshu Harbola}
\author[1]{Anupam Purwar}

\affil[1]{\textbf{AIGuruKul Foundation}}

\maketitle

\maketitle

\footnotetext[1]{Both authors have equal contribution.\\
\href{https://www.linkedin.com/in/chitranshu-harbola/}{Chitranshu Harbola – LinkedIn}, Chennai, India.\\
\href{https://anupam-purwar.github.io/page/}{Anupam Purwar – Webpage}, Delhi, India.}

\begin{abstract}
In the evolving landscape of conversational AI, generating concise, context-aware, and human-like dialogue using small and medium-sized language models (LLMs) remains a complex challenge. KnowSLM introduces a unique framework
to help evaluate enhancement of Small and Medium Language Models with domain-specific knowledge and stylistically adaptive conversational abilities through different techniques viz. Fine-tuning and retrieval-augmented generation (RAG). The \textbf{KnowSLM} framework showcases the complementary benefits of fine-tuning and RAG, where fine-tuning enhances coherence and stylistic consistency within familiar contexts, and RAG effectively injects external or novel information. KnowSLM framework has been applied on a Medium language model, which has been fine-tuned on various synthetic datasets comprising of region-specific food suggestions, poetic response styles, cultural contexts such as MahaKumbh, and recent events such as ISRO’s space docking mission. Our study investigates the influence of LoRA rank, dataset scale, and prompt prefix design on both knowledge retention and stylistic alignment. Although fine-tuning improves fluency and enables stylistic customization, its ability to integrate unseen knowledge is restricted, particularly with smaller datasets. In contrast, RAG-augmented models, equipped to incorporate external documents at inference, demonstrated superior factual accuracy on out-of-distribution prompts, though they lacked the stylistic consistency achieved by fine-tuning. Evaluations by LLM-based judges on knowledge accuracy, conversational quality, and conciseness suggest that fine-tuning is best suited for tone adaptation, whereas RAG excels at real-time knowledge enhancement. Together, these insights inform practical strategies for deploying small and medium LLMs in scalable, human-centric dialogue systems.

\end{abstract}
\maketitle
\section{Introduction}
Large language models have become better and better in conversational style and their knowledge with time by incorporating every growing  information through every cycle of pre-training and launch of new version. Recent research explores how LLMs can be effectively used for dialogue evaluation by leveraging chain-of-thought reasoning, in-context examples, and fine-tuning \cite{jia2024leveraging} \cite{shi2024continual}. However, small and medium size language models viz. LLama 3,  Mistral have not achieved similar performance leaving it to fine tuning or RAG based augmentation with sophisticated multi-shot prompts to have a natural conversation with real world awareness of the facts. Small Language Models (SLMs) are increasingly favored over LLMs for their efficiency, lower cost, and domain-specific adaptability, particularly in resource-constrained environments \cite{wang2024comprehensive}. However, a comprehensive analysis of fine tuning vs RAG to compare their performance on small/medium language models is required to understand the benefits and limitations of both approaches. The current study proposes a framework for the same and also demonstrates the implementation of this framework on LLama 3.3. Besides, the paper also  provides a deep dive into how to generate data for effective fine tuning of language models.

Fine-tuned medium language models(MLMs), like llama 3.3 70B, are highly effective for downstream task due to their efficient resource consumption. This paper employs Low-Rank Adaptation (LoRA), a fine-tuning technique that minimizes the number of trainable parameters by incorporating trainable rank decomposition matrices. This approach allows for efficient and lightweight adaptation of large language models while preserving their performance capabilities \cite{hu2021lora}.

Comprehensive overview of Retrieval-Augmented Generation (RAG), its evolution, and enterprise applicability, serves as the foundation for understanding the retrieval pipeline used in this study.\cite{gautam2024efficacy}. Advancements in retrieval frameworks, such as \textbf{Keyword Augmented Retrieval}, demonstrate the potential of leveraging smaller language models for efficient context discovery, thereby reducing inference costs and response times \cite{purwar2023keyword}. Hybrid retrieval techniques, such as \textbf{COS-Mix}, which integrate cosine similarity and distance measures, have demonstrated improved retrieval performance, particularly in sparse data scenarios \cite{juvekar2024cosmix}. Research on hyperparameters in Retrieval-Augmented Generation (RAG) systems highlights the impact of \textbf{ Context Window Utilization} on retrieval effectiveness and response quality \cite{juvekar2024context}. This insight is relevant to our approach in optimizing LoRA fine-tuning, ensuring that model-generated responses retain high contextual accuracy and coherence. Recent exploration in fine-tuned medium language models (MLMs), such as LLaMA3.3-70B, demonstrate their effectiveness in downstream tasks through parameter-efficient optimized methods such as low rank adaptation (LoRA) and synthetic data generation \cite{xiao2024cora}  \cite{zheng2024llamafactory}. These models balance computational efficiency with high performance using techniques that reduce resource consumption while maintaining or improving accuracy compared to their larger counterparts \cite{kim2024ralora}. In the following, we synthesize insights from recent studies to outline key methodologies to enhance MLM outputs in human-like conversational styles.

The integration of Generative AI with cultural learning systems, as demonstrated in CultureVo’s ICLS framework, highlights the potential of LLM in adaptive content generation and real-time learning assessment \cite{agarwala2024culturevo}. These insights reinforce our approach to crafting instruction prompts that generate responses that reflect different styles and tones of conversation.
VidyaRANG introduces a privacy-preserving conversational learning platform that enhances traditional educational methods by integrating knowledge-augmented retrieval techniques and Large Language Models (LLMs) \cite{harbola2024vidyarang}. Unlike conventional search engines that return an overwhelming list of results, this platform delivers authoritative, context-specific answers—particularly valuable when dealing with organization-specific knowledge that is not publicly accessible. It also enhances scalability by automating knowledge retrieval, enabling instructors to effectively reach and train larger audiences. This aligns with the goals of our study, where we demonstrate that fine-tuning medium-sized LMs, combined with retrieval-augmented strategies, can generate highly relevant and stylistically diverse conversational responses. By tailoring outputs to reflect distinct interpersonal tones—such as those used by various teacher personas—our framework supports more personalized and accurate communication. Ultimately, this helps the system produce not only factually grounded responses but also contextually appropriate and human-like interactions that are more aligned with the needs of specific users. From these set of experiments we aim to extract knowledge through responses generated from the fine-tuned MLM in desired human-like conversational dialogues. This framework creates synthetic dialogues from diverse sources and fine-tune Llama 3.3 70B using optimized LoRA parameters to generate naturally distinct, human-like responses.

\section{Methodology}
\begin{figure}[h]  
    \centering  
    \includegraphics[width=\textwidth, height=0.8\textheight, keepaspectratio]{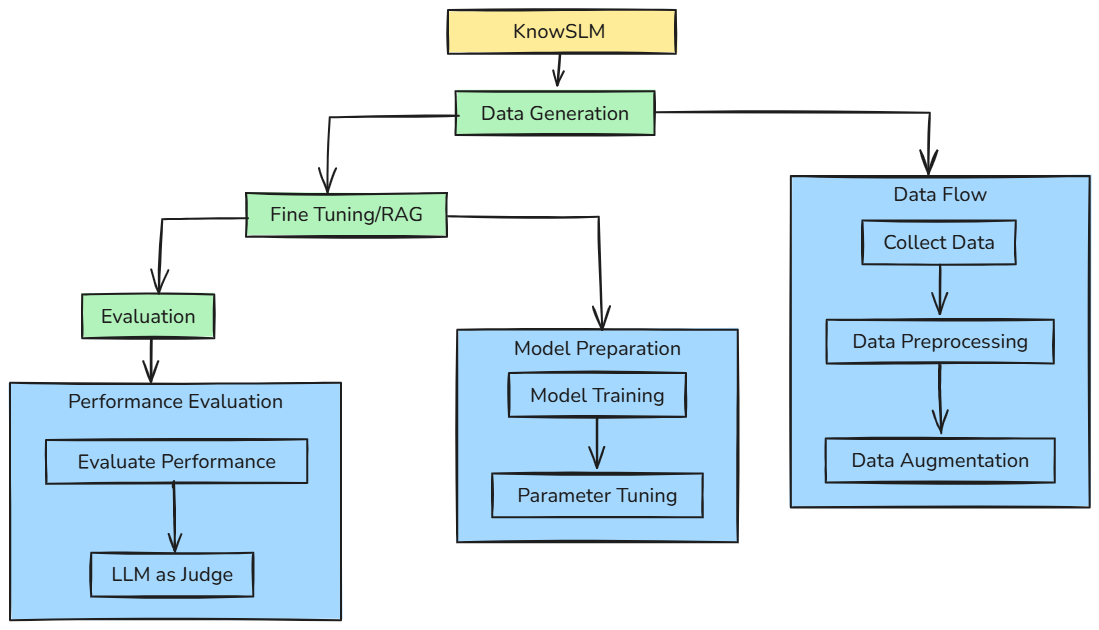}  
    \caption{KnowSLM Framework: proposed methodology consisting of knowledge generation, knowledge augmentation (fine tuning and RAG) and augmented LM evaluation.}  
    \label{fig:datageneration_pipeline}  
\end{figure}

Recent advancements in language modeling (LM) have improved human-like conversational responses from LLM, but challenges remain in maintaining context, coherence, and relevance, particularly in multi-turn conversations, which involves followup questions, proactive messaging. LLM can't produce reliable responses for information which it has never been exposed.To overcome the limitations of LLMs in handling knowledge-intensive tasks, this paper proposes \textbf{KnowTuning}, a fine-tuning approach designed to enhance both fine-grained and coarse-grained knowledge awareness, thereby improving the model's ability to generate contextually rich and accurate conversational outputs.  \cite{lyu2024knowtuning}.

However the responses will still be similar to the parent LLM which does not reflect the follow up questions to carry forward the conversation. We tend to find whether we can add the human-like conversational style and tone in the responses while incorporating knowledge. \textbf{DSPy} explores the limitations of existing language model (LM) pipelines, which typically rely on hard-coded prompt templates that are discovered through trial and error\cite{Khattab2023DSPy}.
DSpy enhances the generation of coherent and contextually relevant dialogues with a focus on question-answer patterns. By refining prompt-output interactions, it addresses issues like prompt misinterpretation and context loss, improving the efficiency and effectiveness of conversational AI systems. We utilize DSPy within a Retrieval-Augmented Generation (RAG) setup to generate human-like interactions from a given knowledge source. The process begins by uploading a knowledge document, from which GPT-4o is prompted—via DSPy—to generate synthetic questions that a human might naturally ask based on the content. These questions are then paired with the relevant knowledge and passed through a second prompt, instructing GPT-4o to generate responses that are not only accurate but also conversational, emotionally aware, contextually grounded, and concise, often including appropriate follow-up questions to maintain dialogue flow.
   
\subsection{Conversational Dialogue Dataset Generation}

This section focuses on creating a conversational dataset for finetuning open source Llama 70B Instruct, Medium Language Model. The entire process is based on the automated extraction, separation, and transformation of textual data from an interview conversation or an article into a structured format suitable for machine learning models. The methodology involves two key steps:\textbf{ data separation} and \textbf{conversation generation}. Below is a detailed breakdown of the process.

\begin{enumerate}  
    \item Data Separation: The first step is to separate the conversational data into distinct components one for the user and one for the assistant. This is essential for creating labeled data, which is a key requirement for training conversational models.
    \item Conversation Generation: Once the data is separated, the next step is to create training examples for finetuning. The goal here is to structure the conversation into a specific format that can be fed into the model for training. Each conversational pair (question and response) will be formatted as a set of roles that a conversational AI would understand.
\end{enumerate}
   \begin{figure}[h]  
    \centering  
    \includegraphics[width=\textwidth, height=0.8\textheight, keepaspectratio]{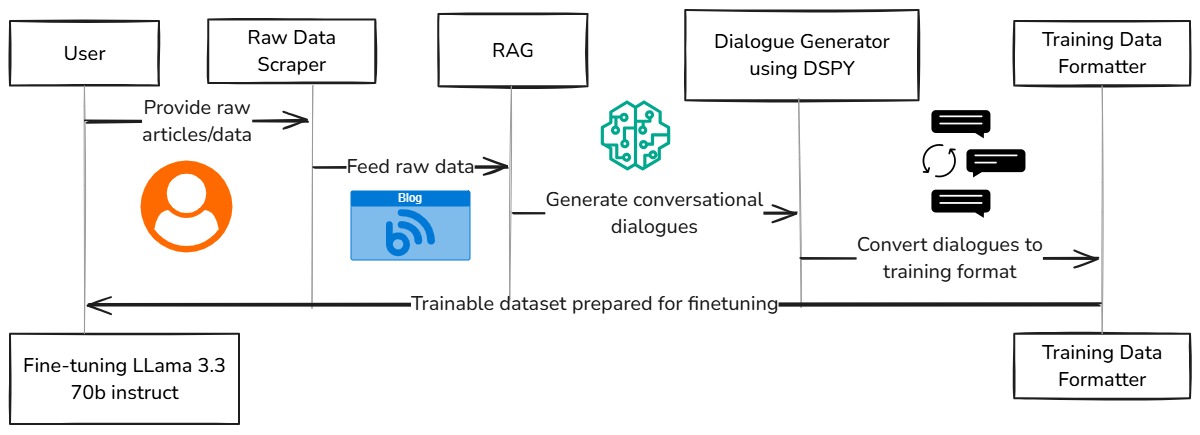}  
    \caption{Synthetic dialogue generation pipeline}  
    \label{fig:datageneration_pipeline}  
\end{figure}  

This methodology provides a streamlined approach for transforming raw informational data into a
structured format that can be used for training LLMs, particularly in training models for natural language understanding and incorporating fact based knowledge, that has not already been exposed to LLM. The goal is to create engaging and natural conversation prompts (questions) and responses (answers) based on the input data. The output will be a CSV file with engaging question-answer pairs related to specific search queries. The questions and answers are designed to simulate natural conversation and may
be used for various purposes like chatbot development, content generation, or data analysis.

Identified Issues in Data Generation:
\begin{itemize}
    \item The word frequently appears at the beginning of questions eg. 'kya',"what","why", leading to redundancy.
    \item The model performs well for English conversations but struggles with Hinglish conversations.
    \item There is repetition in both question structure and phrasing—many questions start with the word 'How', making them feel repetitive despite being different.
\end{itemize}
Below is the refined prompt, ensuring that the generated conversations are unique and well-structured in both English and Hinglish.

\begin{tcolorbox}[colback=blue!5!white, colframe=blue!75!black, title=DSPy Question Generation Prompt]
    
Generate one conversation initiating statement in English/ Hinglish. Frame the conversation in a natural tone and use different starting formats of questioning like 'why', 'when', 'where'. Ensure the questions feel engaging and unique. 

\end{tcolorbox}
\begin{tcolorbox}[colback=blue!5!white, colframe=blue!75!black, title=DSPy Answer Generation Prompt]
    
Give an informative response in English in 2 lines. Ask a thoughtful question at the end in English/ Hinglish. Don't generate further question and answer
statements.

\end{tcolorbox}

\subsection{Fine-Tuning Experiments}

To shape the LLM so that it produces responses for our specific use case, we utilize fine-tuning and optimize over its parameters by providing it the dataset that has enough knowledge and response style from which LLM can learn to structure further responses. In their study on instruction fine-tuning, Zhou et al. \cite{zhou2023lima} present LIMA, a large language model trained with just 1,000 carefully curated prompts and responses. Their findings suggest that most of the knowledge in large language models is learned during pretraining, and minimal instruction tuning is sufficient to produce high-quality outputs. LIMA's performance is comparable or even superior to models trained with extensive reinforcement learning or human feedback, demonstrating the power of efficient fine-tuning. Fine-tuning using LoRA in simpler terms involves training of LLM over few parameters due to which knowledge incorporation gets cost-effective and less time-consuming.
In this paper, we aim to conclude the knowledge incorporation in the fine-tuned LLM and evaluating the responses by LLM-Judge which has been used to rank the responses from the base model and fine-tuned model.
\subsection{Experiment : Llama 3.3 70B Out-of-Scope Knowledge}
In the first experiment, we focused on incorporating the local insights into the LLM where we worked with a synthetic dataset of 7556 dialogues focusing on food and café recommendations in Delhi. When prompted for recommendations or opinions, the names of restaurants and cafes often appeared more frequently in the responses of the fine-tuned model which were present in the training data Table~\ref{tab:model_comparison}. Sometimes the responses overshadowed relevance to the user's query by suggesting a bias toward high-frequency entities in the dataset. Ideally, the model should suggest restaurants that align more accurately with the query's intent.

Additionally, model fine-tuned on dataset structure, paired input-output dialogues in a single language without explicit instructions, the model responded in the same language as the user's input. This design choice led the model to naturally mimic the language of the input query in its output.

\begin{table}[ht]
    \centering
    \resizebox{\textwidth}{!}{%
    \begin{tabular}{|c|c|c|c|c|c|c|}
        \hline
        \textbf{Sno.} & \textbf{Records} & \textbf{Dataset} & \textbf{LoRA r} & \textbf{Alpha} & \textbf{Dropout} & \textbf{Trainable Parameters} \\ 
        \hline
        1 & 7556   & Food Dataset   & 4   & 8   & 0.1   & \(5.18 \times 10^7\)   \\ 
        \hline
        2 & 1300   & Poetry and Food   & 4   & 8   & 0.1   & \(5.18 \times 10^7\)   \\ 
        \hline
        3 & 520    & MahaKumbh   & 8   & 12  & 0.1   & \(1.04 \times 10^8\)  \\ 
        \hline
        4 & 520    & MahaKumbh   & 16  & 10  & 0.1   & \(2.07 \times 10^8\)  \\ 
        \hline
        5 & 520    & MahaKumbh   & 32  & 18  & 0.1   & \(4.14 \times 10^8\)  \\ 
        \hline
        6 & 520    & MahaKumbh   & 64  & 40  & 0.1   & \(8.28 \times 10^8\)  \\ 
        \hline
        7 & 800   & ISRO    & 16  & 10  & 0.1   & \(2.07 \times 10^8\) \\ 
        \hline
        8 & 800   & ISRO   & 8  & 12  & 0.1   & \(1.04 \times 10^8\) \\ 
        \hline
        9 & 7000   & ISRO   & 16  & 10  & 0.1   & \(2.07 \times 10^8\) \\ 
        \hline
    \end{tabular}%
    }
    \caption{This table presents the dataset-specific configurations used to fine-tune LLaMA models with Low-Rank Adaptation (LoRA). The datasets include localized dialogues for food recommendations in Delhi, poetic-cultural content, and domain-specific topics like MahaKumbh and ISRO. Each configuration varies in LoRA rank, alpha, and the number of trainable parameters, reflecting the resource allocation for adapting the model to specific cultural and linguistic contexts. The model trained on the food dataset exhibited strong memorization of entity names, occasionally favoring high-frequency mentions, while naturally mirroring the input language due to its paired dialogue structure.}
    \label{tab:model_comparison}
\end{table}

Integrating Poetic and Domain-Specific Knowledge dataset :
The second experiment aimed to Instruction-tune the model on conversational dialogues to generate distinct responses when prompted with specific instruction.  A synthetic small dataset comprising two subsets was used:

Poetry Dataset: Containing 600 dialogues, each prefixed with the string, "You love to give poetic responses."
Delhi Food Dataset: Containing 800 dialogues, each prefixed with the string, "You have knowledge of Delhi food."
\begin{figure*}[htbp!]
    \centering
    \includegraphics[width = \textwidth]{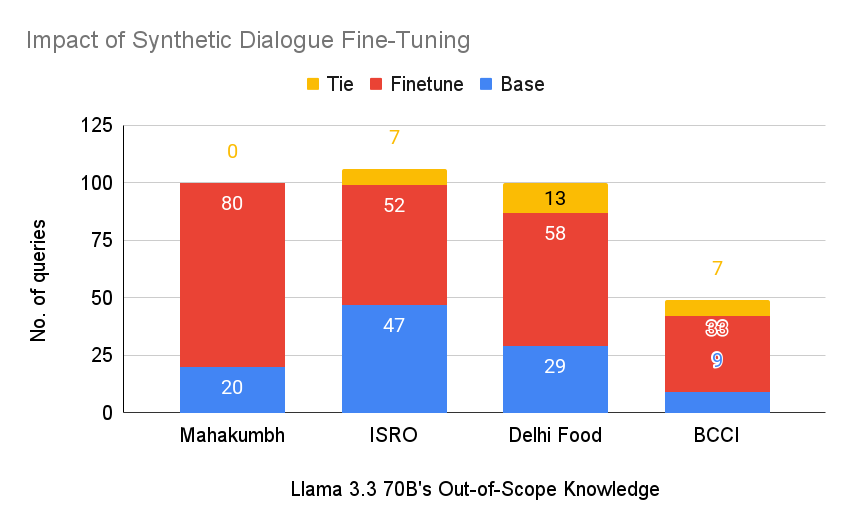}
\caption{Win Chart: Impact of Synthetic Dialogue Fine-Tuning on Unseen Data Knowledge Recall in LLaMA}
    \label{fig: Comparision on Various Knowledge Source}
\end{figure*}1

To adapt fine-tuned model's responses based on prompt's prefix, the model was trained on these combined datasets. For instance, the model generated answers with a poetic tone when the string "You love to give poetic responses" was included in the input. Similarly, the prefix "You have knowledge of Delhi food" guided the model to provide detailed and relevant responses related to Delhi's food scene. 
As shown in Figure~\ref{fig: Comparision on Various Knowledge Source} this approach demonstrated the model's ability to align its response style and tone with the intent specified in the prompt, highlighting the effectiveness of targeted prefix-based fine-tuning.

\subsection{Limitations of Fine-Tuning on Small Datasets}

\begin{figure*}[htbp!]
    \centering
    \includegraphics[width = \textwidth]{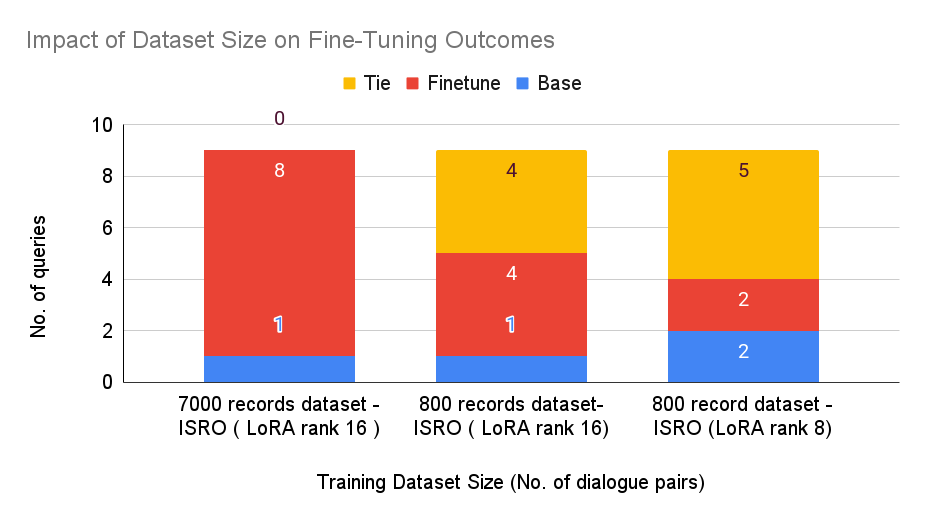}
    
\caption{Win Chart: Evaluating the Impact of Small Dataset Size and Token Length on Fine-Tuning Performance}
    \label{fig: Dataset Token Length Comparision}
\end{figure*}

For experiments conducted with smaller datasets (500–2,000 records), increasing the LoRA parameters r and alpha up to 64 and 40, respectively, did not yield results better than the base model. The fine-tuned model demonstrated an ability to adopt the desired tone in its responses, reflecting the poetic style intended. However, as shown in Figure~\ref{fig: Dataset Token Length Comparision} it struggled to generate exact phrases as expected, indicating limited learning capacity from the small dataset. 

The base model demonstrated pre-existing knowledge; however, its responses were often misaligned with the specific queries, limiting the observable improvements from fine-tuning. Furthermore, evaluations using an LLM-based judge did not reveal any significant performance gains in the fine-tuned model. These results suggest that fine-tuning on smaller datasets, even with data replication and adjusting LoRA parameter values, may have limited effectiveness in outperforming the base model, especially when the base model already has some prior knowledge of the subject.

 The impact of dataset size on model performance was evaluated. Responses generated from the 7000-record dataset demonstrated significant improvements over those produced by the 800-record dataset. With the 800-record dataset, the generated responses did not show considerable enhancement. While increasing the number of training epochs could potentially improve the performance, it would also extend the training time. When comparing different LoRA configurations, there was minimal performance variation with the 800-record dataset, likely due to the small dataset size and limited training epochs. Fine-tuning with the 7000-record dataset took approximately 28 minutes, about 10 times longer than the 3 minutes required for the 800-record dataset. Additionally, the fine-tuned model trained on the larger dataset showed a tendency to incorporate words that appeared more frequently in the training data, influencing the vocabulary used in the generated responses.

 \subsection{Evaluation of Fine-Tuned and RAG-Augmented Models}

To evaluate the impact of incorporating external knowledge into a fine-tuned model, we conducted a comparative study involving two approaches: (1) a fine-tuned model on synthetic conversational dialogues derived from concise knowledge inputs, and (2) a medium language model that receives an external knowledge document during inference.

\begin{figure}[h]  
    \centering  
    \includegraphics[width=\textwidth, height=0.8\textheight, keepaspectratio]{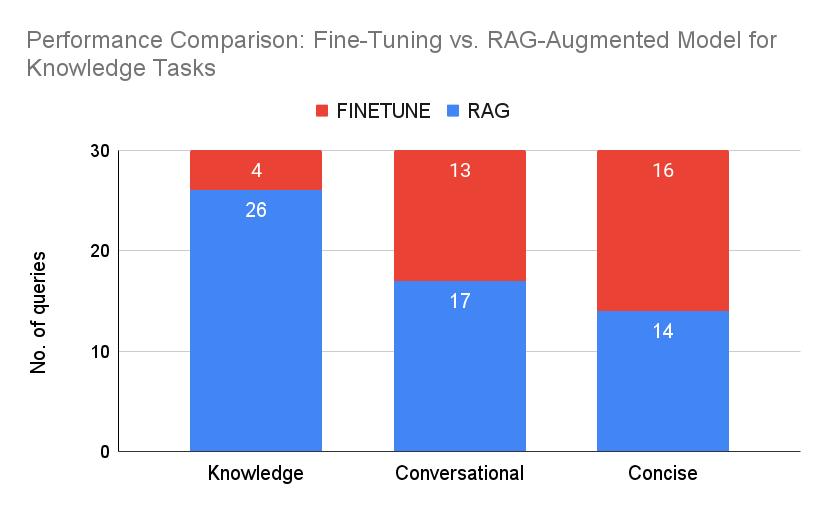}  
    \caption{Win Chart: This figure illustrates the comparative performance of a fine-tuned LLaMA 3.3 70B model and a RAG-augmented variant on unseen knowledge-intensive queries.}  
    \label{fig:Comparision RAG vs Finetune}  
\end{figure}

The base model used for this study was the LLaMA 3.3 70B Instruct, which lacked the domain-specific knowledge present in the evaluation set. To test the models' ability to respond to queries requiring external information beyond its pre-training data, we curated 30 prompts targeting content unavailable in the base model. Both the fine-tuned and RAG-augmented models were tasked with generating responses to these prompts.

The generated responses were then evaluated by an LLM-based judge using three criteria: Knowledge, Conversational Quality, and Conciseness. As illustrated in Figure~\ref{fig:Comparision RAG vs Finetune}, the RAG model outperformed the fine-tuned-only model in terms of knowledge accuracy, demonstrating the effectiveness of integrating external documents during inference. While using RAG, the prompts were specifically crafted to generate conversational and concise responses by utilizing the context provided in the document. Meanwhile, the fine-tuned model achieved moderately strong results in conversational tone and conciseness, indicating that even with limited data and compute resources, fine-tuning can lead to noticeable improvements in human-like dialogue generation.

These experiments underscore the utility of lightweight fine-tuning methods like Low-Rank Adaptation (LoRA), which allow efficient adaptation of medium-sized language models (MLMs) such as LLaMA 3.3 70B. Our approach focuses on generating distinct conversational styles by fine-tuning on synthetic dialogues created from structured knowledge sources such as articles, books, and current events.

\subsection{Impact of LoRA Rank on Fine-Tuning Performance}
To further investigate the effect of Low-Rank Adaptation (LoRA) on fine-tuning performance, we conducted an experiment evaluating four different LoRA rank values: 4, 6, 8, and 32. The objective was to assess how varying the LoRA rank influences knowledge incorporation and response quality when fine-tuning LLaMA 3.3 70B on a dataset containing knowledge absent in the base model.
\begin{figure*}[htbp!]
    \centering
    \includegraphics[width = \textwidth]{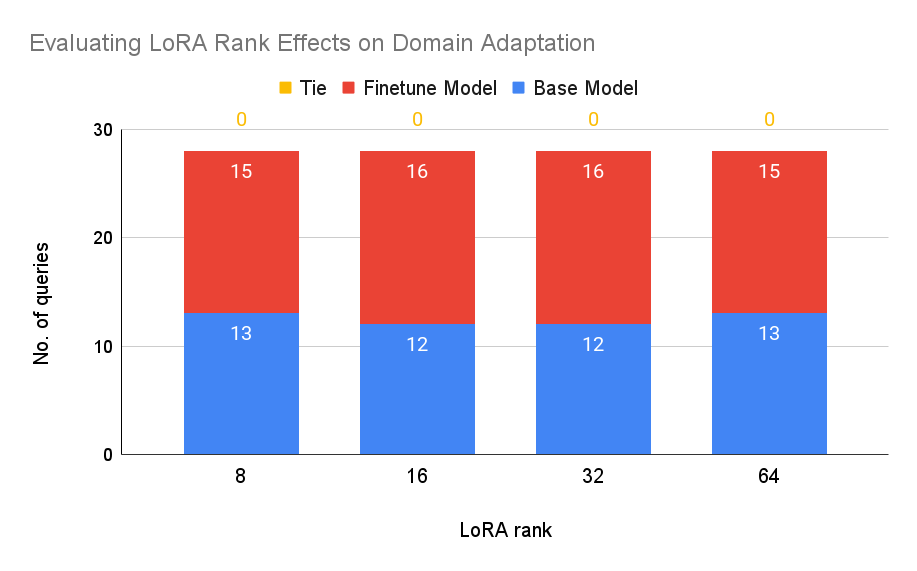}
\caption{Win Chart: Impact of LoRA Rank on Knowledge Infusion and Response Quality in LLaMA 3.3 70B Fine-Tuning}
    \label{fig: 
Comparision lora parameters}
\end{figure*} 

Given the constraints of a limited dataset and compute resources, the results across different ranks remained relatively similar on our test set. However, prior research~\cite{ed-dib2024gelora} suggests that LoRA rank tuning plays a crucial role in optimizing trade-offs between computational efficiency and model expressiveness: These configurations are generally more computationally efficient and are suitable for lightweight domain adaptation tasks where subtle knowledge infusions are sufficient \cite{hu2021lora}. However, they may struggle to fully capture complex knowledge representations. Studies suggest that rank 8 strikes a balance between efficiency and expressiveness, making it effective for conversational fine-tuning while maintaining fast inference speeds \cite{xiao2023scaling}. A higher LoRA rank allows the model to retain more fine-grained knowledge, which is beneficial for technical and domain-specific tasks that require deeper contextual understanding \cite{dettmers2023qlora}.

\section{Results and Discussion}

\begin{figure}[h]  
    \centering  
    \includegraphics[width=\textwidth, height=0.8\textheight, keepaspectratio]{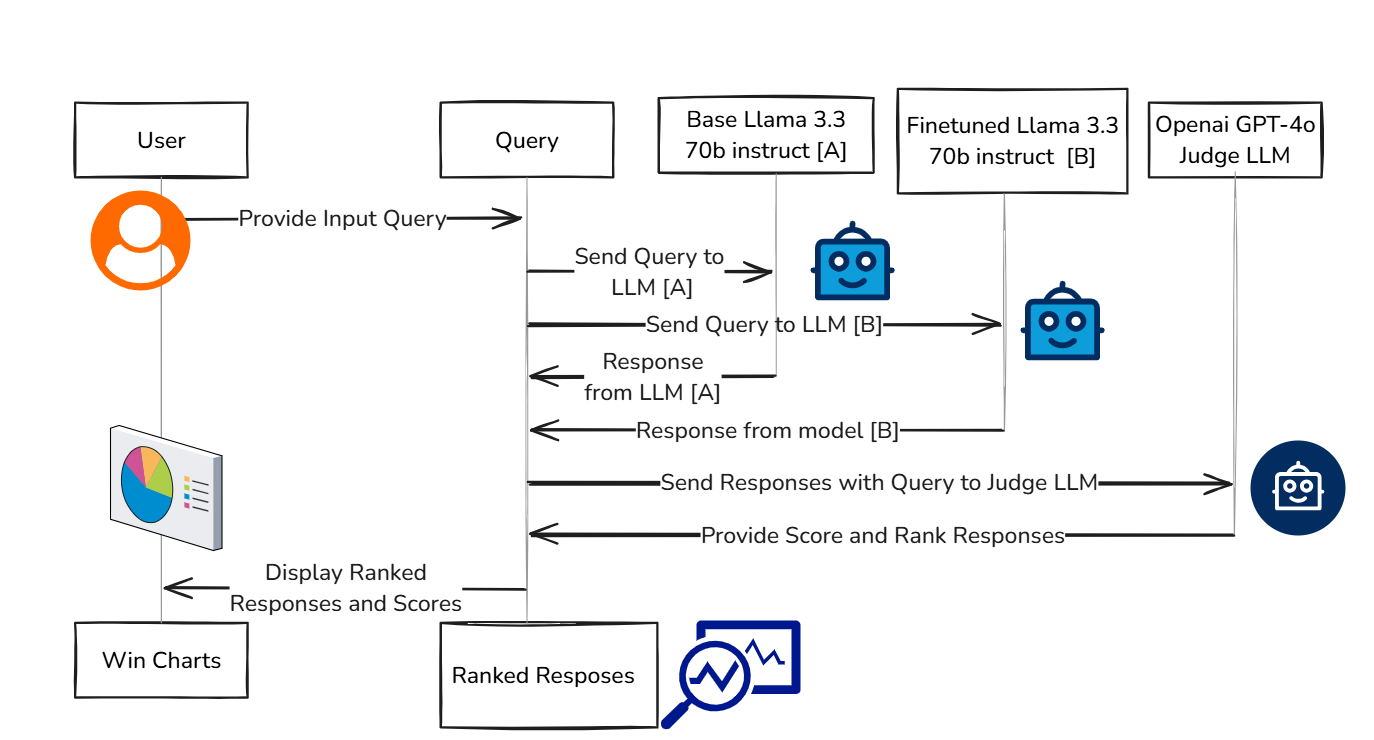}  
    \caption{Flow diagram illustrating how responses from the base LLaMA 3.3 70B model and the fine-tuned LLaMA 3.3 70B model, generated for the same input query, are evaluated by the Judge GPT (GPT-4o). The judge ranks these responses based on quality, which is then used to generate comparative performance charts.}

    \label{fig:llm judge workflow}  
\end{figure}

Evaluations using an LLM-based judge highlighted the fine-tuned models' ability to respond naturally but revealed limitations in exact knowledge recall. In their exploration of knowledge integration in large language models (LLMs),  \cite{Gu2024LLMJudge} provide a comprehensive survey on the use of LLMs as evaluative agents in complex tasks. Their study underscores the potential of LLMs to serve as reliable judges by enhancing evaluative consistency, reducing bias, and managing subjectivity in assessments. The authors stress the importance of developing rigorous evaluation protocols to establish trustworthy 'LLM-as-Judge' frameworks. As illustrated in Figure~\ref{fig:llm judge workflow}, models like GPT-4o can effectively evaluate and rank responses from both base and fine-tuned models when given the same prompt. However, such evaluators may still exhibit positional bias—tending to prefer responses presented earlier, regardless of their actual quality.

\begin{tcolorbox}[colback=blue!5!white, colframe=blue!75!black, title=RAG based LLM judge prompt]

Please act as an impartial judge and evaluate the quality of the response provided by two AI assistants to the input prompt.
The responses should reflect knowledge of \textit{{KNOWLEDGE SOURCE}} demonstrating specific and knowledgeable insights from\textit{ {CONTEXT} }about the query.
Avoid positional Biasness.
Just declare which response is better and provide one statement why.
        User's Query:\textit{ PROMPT}
        Assistant A Response: \textit{BASE MODEL RESPONSE}
        Assistant B Response: \textit{FINE-TUNED MODEL RESPONSE}
        You should choose the assistant that produces a better generation. Avoid positional biases and ensure that the order in which the responses were presented does not influence your decision. Be as objective as possible. After providing your explanation, output your final verdict strictly following this format: [[A]] if assistant A is better, [[B]] if assistant B is better, and [[C]] for a tie.

\end{tcolorbox}

Responses from the base and fine-tuned models were evaluated using a RAG-based LLM judge with contextual understanding, which ranked them as better based on quality. In this observation from the Delhi food dataset, for some queries it favored the base model's response as it directly addressed the topic of rooftop dining by mentioning a specific café in Delhi. In contrast, the fine-tuned model discussed a restaurant without relating it to rooftop dining, making the response less relevant to the prompt.

\begin{table}[h]
    \centering
    \resizebox{\textwidth}{!}{%
    \begin{tabular}{|c|c|c|c|c|}
        \hline
        \textbf{LoRA Rank ($r$)} & \textbf{LoRA Alpha ($\alpha$)} & \textbf{Trainable Params} & \textbf{Total Params} & \textbf{Percentage} \\
        \hline
        2  & 2  & \(2.59 \times 10^7\)  & \(7.06 \times 10^{10}\)  & \(3.67 \times 10^{-2}\) \\
        4  & 2  & \(5.18 \times 10^7\)  & \(7.06 \times 10^{10}\)  & \(7.33 \times 10^{-2}\) \\
        8  & 2  & \(1.04 \times 10^8\)  & \(7.07 \times 10^{10}\)  & \(1.47 \times 10^{-1}\) \\
        16 & 2  & \(2.07 \times 10^8\)  & \(7.08 \times 10^{10}\)  & \(2.93 \times 10^{-1}\) \\
        \hline
        2  & 4  & \(2.59 \times 10^7\)  & \(7.06 \times 10^{10}\)  & \(3.67 \times 10^{-2}\) \\
        4  & 4  & \(5.18 \times 10^7\)  & \(7.06 \times 10^{10}\)  & \(7.33 \times 10^{-2}\) \\
        8  & 4  & \(1.04 \times 10^8\)  & \(7.07 \times 10^{10}\)  & \(1.47 \times 10^{-1}\) \\
        16 & 4  & \(2.07 \times 10^8\)  & \(7.08 \times 10^{10}\)  & \(2.93 \times 10^{-1}\) \\
        \hline
    \end{tabular}%
    }
    \caption{The table presents the impact of varying hyperparameters on the fraction of trainable parameters with respect to the total number of parameters in the model.}
    \label{tab:lora_params}
\end{table}

\begin{figure*}[htbp!]
    \centering
    \includegraphics[width = \textwidth]{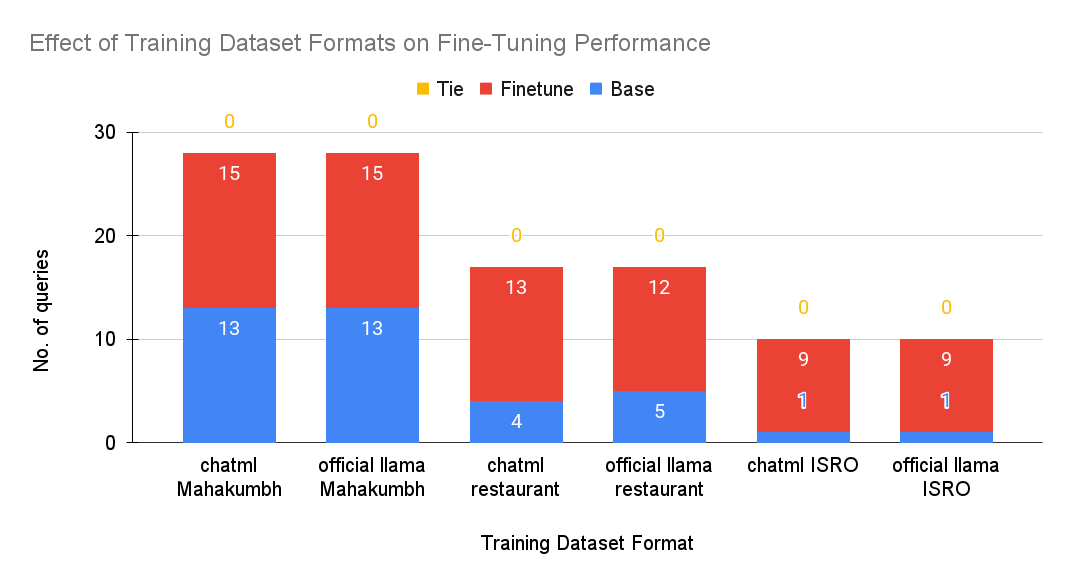}
\caption{Win Chart: Comparative Analysis of Dataset Formats for LLaMA 3.3 70B Fine-Tuning}
    \label{fig: FINETUNING DATASET COMPARISION}
\end{figure*}

Developing conversational dataset relies on other LLM and knowledge sources which are the bottleneck to build the robust model. The performance of the model is directly proportional to the quality and quantity of the dataset. For different ranges of LoRA ranks, LLM-Judge was not able to differentiate well with a small dataset of less than 100k token.
However, the performance increased by 4 times as we got the dataset above a certain range, which also consumed heavy  GPU resources and increased training time.
We utilized a serverless cloud platform for model fine-tuning, where we used H100 GPU, for which we noted the average cost per second to be 0.0022\$ refer Table~\ref{tab:dataset_cost}.

\begin{table}[h]
    \centering
    \resizebox{\textwidth}{!}{%
    \begin{tabular}{|r|r|r|r|r|r|}
        \hline
         \textbf{Dataset Length } & \textbf{Training Time (s)} & \textbf{Cost (\$)} & \textbf{Cost/Second (\$)} & \textbf{Tokens per Second} \\
        \hline
         91,273   & 1,285  & \(2.25 \times 10^0\)  & \(1.75 \times 10^{-3}\)  & \(7.005 \times 10^1\) \\
         184,949  & 2,149  & \(5.6 \times 10^0\)   & \(2.6 \times 10^{-3}\)   & \(8.606 \times 10^1\) \\
         33,644   & 462    & \(1.3 \times 10^0\)   & \(2.814 \times 10^{-3}\) & \(7.282 \times 10^1\) \\
         509,634 & 8,012  & \(1.355 \times 10^1\)  & \(1.691 \times 10^{-3}\) & \(6.361 \times 10^1\) \\
         271,989  & 2,852  & \(7.1 \times 10^0\)   & \(2.48 \times 10^{-3}\)  & \(9.559 \times 10^1\) \\
        \hline
    \end{tabular}%
    }
    \caption{This table highlights the relationship between Dataset length/size(no. of tokens), training time, cost, and throughput. As dataset size increases, both training time and cost scale significantly, emphasizing the resource demands of high-quality fine-tuning.}
    \label{tab:dataset_cost}
\end{table}
Selecting the appropriate LoRA rank is challenging, as its performance varies across different dataset sizes, with the rank determining the number of trainable parameters and the alpha value modulating its influence. The cost of generating a synthetic dataset and evaluation using LLM-Judge is an additional cost. Moreover, achieving an optimal balance between model performance and computational efficiency requires extensive experimentation across different configurations.

\section{Conclusion}
\textbf{KnowSLM} presents an approach to enrich small and medium-sized language models with external knowledge and concise, human-like conversational responses. It demonstrates the effective use of retrieval-augmented generation (RAG) for dynamic knowledge integration, alongside fine-tuning on synthetically generated dialogue pairs derived from source datasets to infuse stylistic and tonal authenticity. Our experiments underscore the nuanced trade-offs involved in enhancing LLaMA 3.3 70B through fine-tuning and retrieval-augmented strategies. We observed that local and stylistic adaptations proved effective in aligning model outputs with cultural and tonal expectations, the benefits were contingent on both data quality and scale. RAG-based augmentation offered clear advantages in factual accuracy, on the other hand fine-tuning better preserved conversational fluency. However, fine-tuning on small datasets yielded limited gains, with significant improvements only emerging at larger scales—albeit with increased computational cost. These findings highlight the critical balance between dataset size, model performance, and resource efficiency in building robust domain-specific language models.
\begin{itemize}
    \item \textbf{Enhancing LLaMA 3.3 70B with Local and Stylistic Knowledge} Fine-tuning on Delhi-specific food dialogues helped the model generate culturally relevant recommendations, though it sometimes favored frequently mentioned entities.
    
    \item \textbf{Evaluation of Fine-Tuned and RAG-Augmented Models}
The RAG-augmented model showed superior knowledge accuracy by leveraging external documents at inference, while the fine-tuned model maintained better conversational tone and conciseness.
    \item \textbf{Limitations of Fine-Tuning on Small Datasets}
    Fine-tuning on datasets with fewer than 100k dataset tokens showed minimal improvement over the base model, even with increased LoRA parameters, due to limited learning capacity and the base model's pre-existing knowledge.
    Performance gains became noticeable only with larger datasets.
    \item \textbf{Scaling Costs and Performance in Fine-Tuning}
Larger datasets led to exponential performance gains but at the cost of higher GPU usage and training expenses, highlighting the trade-off between fine-tuning effectiveness.
\end{itemize}

\section*{Acknowledgements}

We would like to express our sincere gratitude to \textbf{Chitraksh Chavan}, \textbf{Mehul Kohad}, \textbf{Rahul Sundar}, and \textbf{Ujjwal Bokde} for their valuable support throughout the course of this work.


\bibliography{main_AIMLSys}

\end{document}